\documentclass[letterpaper, 10 pt, conference, preprint]{ieeeconf}  

\IEEEoverridecommandlockouts                              

\usepackage{csquotes}
\usepackage{caption}
\usepackage{subcaption}
\usepackage{graphicx}
\usepackage{notoccite}
\usepackage{amsmath}
\usepackage{tabularx}
\usepackage[acronym]{glossaries}
\usepackage{amsfonts}
\usepackage{hyperref}
\usepackage{url}
\usepackage{xcolor}
\usepackage{siunitx}
\usepackage{array,makecell}
\usepackage{multirow}
\usepackage{multicol}
\usepackage{adjustbox}

\usepackage{pgfplots}
\pgfplotsset{compat=newest}
\usepgfplotslibrary{groupplots}
\usepgfplotslibrary{dateplot}
\usepackage{tikz}
\usetikzlibrary{patterns}

\title{\LARGE \bf
On Learning the Tail Quantiles of Driving Behavior Distributions via Quantile Regression and Flows}

\author{Jia Yu Tee$^{1,2}$, Oliver De Candido$^{1}$, Wolfgang Utschick$^{1}$, and Philipp Geiger$^{2}$
\thanks{$^{1}$Chair for Methods of Signal Processing (MSV),
        Technical University of Munich, 80333 Munich, Germany. Correspondence to:
        {\tt\small jiayu.tee@tum.de}.}%
\thanks{$^{2}$Bosch Center for Artificial Intelligenc, 71272 Renningen, Germany.}%
}

\makenoidxglossaries

\newacronym{ad}{AD}{Autonomous Driving}
\newacronym{ade}{ADE}{Average Displacement Error}
\newacronym{fde}{FDE}{Final Displacement Error}
\newacronym{qfde}{QFDE}{Quantile Final Displacement Error}
\newacronym{idm}{IDM}{Intelligent Driver Model}
\newacronym{gail}{GAIL}{Generative Adversarial Imitation Learning}
\newacronym{fagil}{FAGIL}{Fail-Safe Generative Adversarial Imitation Learning}
\newacronym{mlp}{MLP}{Multi-Layer Perceptron}
\newacronym{mlps}{MLPs}{Multi-Layer Perceptrons}
\newacronym{nf}{NF}{Normalizing Flow}
\newacronym{vaes}{VAEs}{Variational Autoencoders}
\newacronym{gans}{GANs}{Generative Adversarial Networks}
\newacronym{nn}{NN}{Neural Network}
\newacronym{qfs}{QFs}{Quantile Flows}
\newacronym{ols}{OLS}{Ordinary Least Squares}
\newacronym{ls}{LS}{Least Squares}
\newacronym{qr}{QR}{Quantile Regression}
\newacronym{qrnn}{QRNN}{Quantile Regression Neural Network}
\newacronym{dbn}{DBN}{Dynamic Bayeisan Network}
\newacronym{hmm}{HMM}{Hidden Markov Model}
\newacronym{1d}{1D}{one-dimensional}
\newacronym{2d}{2D}{two-dimensional}
\newacronym{cdf}{CDF}{Cumulative Distribution Function}
\newacronym{tal}{TAL}{Tilted Absolute Loss}
\newacronym{nl}{NL}{Nonlinear}
\newacronym{nlsq}{NLSQ}{Nonlinear Squared Transformation}
\newacronym{mdp}{MDP}{Markov Decision Process}
\newacronym{mse}{MSE}{Mean Squared Error}
\newacronym{mae}{MAE}{Mean Absolute Error}
\newacronym{rmse}{RMSE}{Root Mean Squared Error}
\newacronym{iid}{i.i.d}{independent and identically distributed}
\newacronym{av}{AV}{Autonomous Vehicle}
\newacronym{dhw}{DHW}{Distance Headway}
\newacronym{thw}{THW}{Time Headway}
\newacronym{ttc}{TTC}{Time-To-Collision}
\newacronym{aqf}{AQF}{Autoregressive Quantile Flow}
\newacronym{anf}{ANF}{Autoregressive Normalizing Flow}
\newacronym{af}{AF}{Autoregerssive Flow}
\newacronym{afs}{AFs}{Autoregressive Flows}
\newacronym{qfr}{QFR}{Quantile Flow Regression}
\newacronym{highd}{highD}{Highway Drone Dataset}
\newacronym{dgp}{DGP}{Diagonal Gaussian Policy}
\newacronym{tpm}{TPM}{Trajectory Prediction Model}
\newacronym{relu}{ReLU}{Rectified Linear Unit}

\newglossaryentry{angelsperarea}{
	name = $a$ ,
	description = The number of angels per unit area,
}
\newglossaryentry{numofangels}{
	name = $N$ ,
	description = The number of angels per needle point
}
\newglossaryentry{areaofneedle}{
	name = $A$ ,
	description = The area of the needle point
}

\begin{document}

\maketitle
\thispagestyle{empty}
\pagestyle{plain} 

\newcommand{\cp}[1]{\textcolor{blue}{#1}}
\begin{abstract}


Towards safe autonomous driving (AD),
we consider the problem of learning models that accurately capture the diversity and tail quantiles of human driver behavior probability distributions, in interaction with an AD vehicle.
Such models, which predict drivers' continuous actions from their states, are particularly relevant for closing the gap between AD agent simulations and reality.
To this end, we adapt two flexible quantile learning frameworks for this setting that avoid strong distributional assumptions:
(1)~quantile regression (based on the titled absolute loss), and 
(2)~autoregressive quantile flows (a version of normalizing flows). 
Training happens in a behavior cloning-fashion.
We use the highD dataset consisting of driver trajectories on several highways.
We evaluate our approach in a one-step acceleration prediction task, and in multi-step driver simulation rollouts.
We report quantitative results using the tilted absolute loss as metric, give qualitative examples showing that realistic extremal behavior can be learned, and discuss the main insights.

\end{abstract}

\section{INTRODUCTION}

\gls{ad} is a rapidly evolving technology that could potentially revolutionize the automotive industry. Despite being a prevalent topic in recent years, the safety of autonomous vehicles and surrounding road users still remains a major concern. It is increasingly common to find reports about crashes involving autonomous vehicles due to rare road situations and unexpected driving behavior of surrounding human drivers. This highlights the fact that 
there are still significant safety challenges that need to be addressed in this field.

For such safety, it is important to correctly predict the behavior of the human drivers surrounding an AD vehicle. This helps in designing AD controllers, as well as for scalably testing a given AD controller through realistic multi-agent simulations \cite{suo2021trafficsim,igl_symphony:_2022}.
But it is not enough to just predict the average or most likely trajectories (i.e., point predictions).
Instead, we need to account for 
the \emph{uncertainty} and \emph{variability} inherent in the traffic behavior \cite{suo2021trafficsim} (in probabilistic terms, ideally we would like to know the probability of collision for a given AD controller).
And we need \emph{metrics} that evaluate how well models capture such variability.



\begin{figure}[t]
	\begin{center}
		\includegraphics[width=\linewidth]{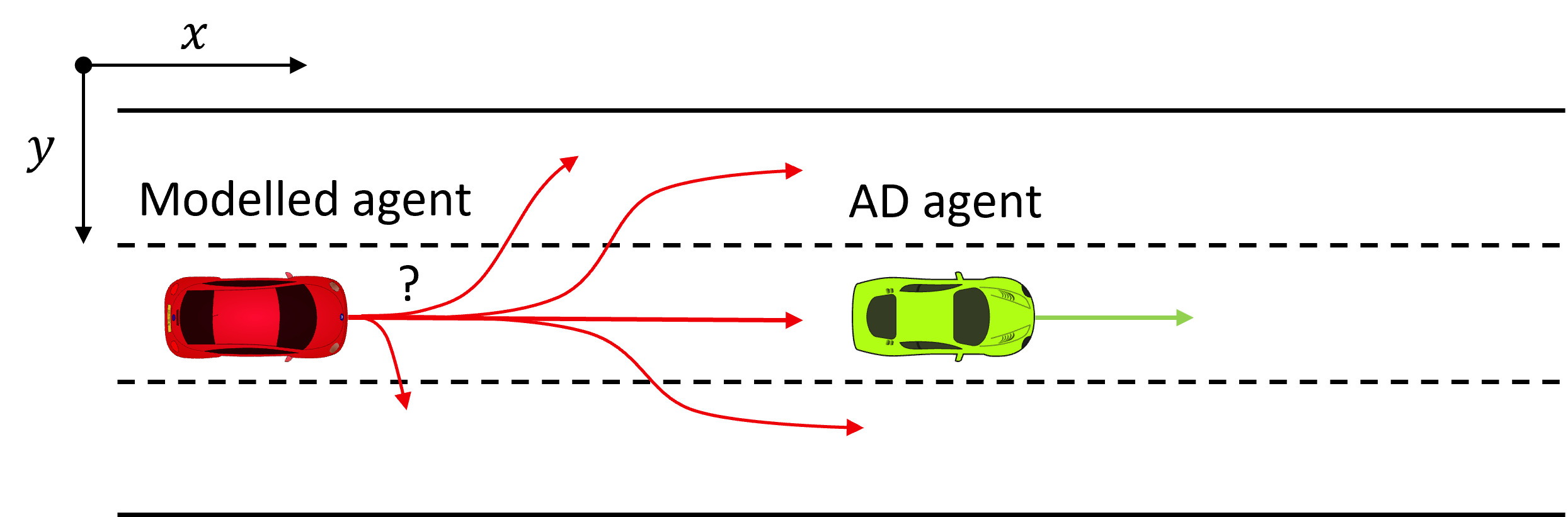}
		\caption{For the green AD agent to assess how hard it can brake without causing a collision, it needs a detailed prediction that captures the diversity of the red modeled agent's realistic responses. 
		\label{fig:uncertaintrajectoryillustration}}
	\end{center}
\end{figure}

Let us give an example that is simple but resembles recent reports of actual AD collisions \cite{dmv,gua}: 
As depicted in Figure~\ref{fig:uncertaintrajectoryillustration}, consider a highway scenario involving two vehicles, the AD ``ego'' vehicle in green, and the surrounding vehicle behind it in red, following rather closely. 
Assume the green AD vehicle has to slow down due to some traffic event in front of it.
This leads to a difficult trade-off, to find the sweet spot between braking hard enough to avoid the front traffic, but also braking gently enough to give the red follower enough space/time to respond safely, if possible.
So, for each of the green AD's braking options, it needs to know how the red car would respond, and if it would lead to a crash.
But there rarely is a single plausible response, instead, there is uncertainty and diversity.
So we need a good prediction of the ``worst realistic'' response of the red car.
(This cannot simply be the physically possible worst case, because this would mean that the red car would even accelerate, making the situation unsolvable.)
Let us say that for a hard brake of the green ego, the \emph{``average''} red driver would manage to safely maneuver and avoid the collision. But it also seems realistic that, at least sometimes, the red driver is in a cognitive state not observant enough of the surrounding and reacts too slowly, leading to a collision as a \emph{rare but still realistic} worst case for this braking option. Similar scenarios also occur at intersections or merging lanes.
%

While this diversity of plausible reactions can generically be captured by a probability distribution over actions, it is crucial to accurately estimate/evaluate \emph{particular aspects of the distribution}, with as little bias as possible. In particular, intuitively, we need to accurately model the \emph{``range''} of realistic interactive driving behaviors.
For this, it is key to know the very high and very low \emph{tail quantiles} of the action distribution, i.e., the extremal actions that are \emph{rare} to happen and distant from the distribution's center, \emph{between which the realistic range lies} (a formal definition of quantiles follows in Section~\ref{subsec:background}).
This leads to our \textbf{overarching objective}, informally expressed as follows: 
\begin{displayquote}
\emph{To learn a flexible model that can reliably predict the tail quantiles of the actions of drivers (that surround and interact with an AD vehicle).}
\end{displayquote} 

\subsection{Related Work}

Extensive research has been conducted on the general problem of predicting human-like driving behaviors \cite{rudenko2020human}, often using machine learning methods.
Some research like \cite{chandra_forecasting_2019, 9195282}, focuses on point forecasts, e.g., mean or median, without considering the associated variability inherent in the traffic.
%
%
%
Some approaches \cite{_probabilistic_????,_probabilistic_2022}, including \gls{gail} \cite{_pdf_????-1},
take probabilistic approaches to address this variability.
However, they assume that the action distribution of other vehicles to be Gaussian, which may not reflect the true data distribution.

Most relevant to the present work, 
recent probabilistic approaches such as \cite{suo2021trafficsim,igl_symphony:_2022,gupta2018social,ma2020normalizing} allow for flexible driver behavior distributions by harnessing variational, generative adversarial, or normalizing flow learning methods.
In particular,
\cite{suo2021trafficsim,igl_symphony:_2022} aim to close the gap between probabilistic closed-loop driver simulations and reality, for simulation-based safety validation. But it is unclear how accurately these models capture the tail quantiles, as this was not explicitly evaluated in these publications.

Beyond point forecasts and probabilistic predictions, there are also some research works that capture the uncertainty about other vehicles based on worst-case reachability analysis, e.g., \cite{9302873,geiger2022failsafe}.
However, such worst-case analysis tends to be over-cautious,
leading to situations where there is no such worst-case safe AD action at all.

\subsection{Main Contributions and Structure of This Work}
The main contributions of our work are summarized as follows:
\begin{enumerate}
	\item We build on the \gls{qr} framework \cite{koenker_quantile_2001}, that uses a dedicated quantile loss function, to estimate \gls{1d} 
	tail events of driver behavior, in particular longitudinal acceleration actions.
	\item We expand the quantile estimation by building on the \gls{aqf} \cite{si_autoregressive_2021} framework, a version of normalizing flows \cite{kobyzev_normalizing_2021,papamakarios2021normalizing}, to learn the \gls{2d} conditional acceleration distribution, covering multiple quantile levels simultaneously.
	\item We conduct experiments using the highD dataset \cite{highDdataset}, for one-step driver predictions and multi-step rollout simulations. We report quantile metric evaluations as well as qualitative results, for the task of diverse driver modeling (Section \ref{sec:experiments}).
\end{enumerate}

It is important to note that the \gls{aqf} approach adapted from \cite{si_autoregressive_2021} is a rather novel method. Therefore, the results presented in this study represent the first steps in using this method for the problem of driver modeling.




It also needs to be emphasized that the methods we build on are for generic quantile estimation, while estimating extreme quantiles is generally difficult, due to the issue of data sparsity and estimator variability for those rare tail events \cite{Wang.2012}.
Our approach cannot fundamentally circumvent this difficulty (future steps to do so may incorporate additional biases or Extreme Value Theory \cite{aasljung2017using}). 
But a key feature of our present approach is that if enough data is available, then it can provide accurate tail quantile estimates, based on flexible models plus the explicit quantile loss formulation. 
In the course of this work,
we will focus on quantile levels like 99\%, as a trade-off between the levels that we can accurately estimate from the given data size, and levels that are needed for eventual AD safety.

This paper is organized as follows:
Section \ref{sec:settingproblem} provides background and problem formulation.
Section \ref{sec:approach} describes how the quantile estimation approaches were adapted and applied to address the core objective of the paper. 
In Section \ref{sec:experiments}, a detailed description of the experimental setup, dataset,
evaluation benchmarks, baselines, and results is given.
Section \ref{sec:discussion} discusses the key findings derived from conducting the experiments, along with suggestions for potential future work. 
Finally, Section \ref{sec:conclusion} provides concluding remarks.


\section{SETTING AND PROBLEM FORMULATION}
\label{sec:settingproblem}

\subsection{Background}
\label{subsec:background}
Throughout the paper,
$P(\cdot)$ denotes a probability distriution, and $P(\cdot|\cdot)$ the conditional distribution.
Intuitively, a quantile is defined as the specific point that divides a probability distribution into two intervals, such that a specific percentage of observations fall below that value. Formally, consider a \gls{1d} random variable $X$ with a \gls{cdf} $F_X: \mathbb{R} \to [0,1]$, $F_X(x) := P(X \leq x)$. The \emph{quantile function} $Q: \alpha \in [0,1] \to \mathbb{R}$ of \emph{quantile level} $\alpha$ returns a threshold value $x$ below which $\alpha \cdot100\%$ of the random samples from the distribution would fall \cite{wasserman2010statistics}. Formally, this is written as
\begin{align}
\label{eq:quantile}
	Q(\alpha) &= \textrm{inf}\{x \in \mathbb{R} : \alpha \leq F_X(x)\}.
\end{align}
In the case where the \gls{cdf} is \emph{strictly monotonic}, the quantile function is simply the inverse $Q=F_X^{-1}$ \cite{_quantiles_????}.
Note that quantiles are generically only defined for the case of 1D distributions.
However, in the case of more than one dimension, one can nonetheless consider the quantile of any one dimension's probability \emph{conditional} on the remaining dimensions \cite{si_autoregressive_2021}.

\subsection{Problem Formulation}
\label{sec:problem}

We consider driving scenarios consisting of a \emph{modeled agent}, i.e., the driver agent we want to predict, and its surrounding, comprising the following variables:
\begin{itemize}
\item \textit{State $s \in S$}: the state of all driving agents considered (in particular, the modeled agent and the AD agent) and environment in a specific situation, where $S$ denotes the set of all possible states. 
\item \textit{Action $a \in A$}: the driving action taken by the modeled agent, where $A$ represents the set of all feasible driving actions. 
\end{itemize}
We assume the state-action pairs $(s_0,a_0),(s_1,a_1),\ldots$ 
are \gls{iid}, 
drawn from some joint distribution $P(a,s)$. In particular, the driving action $a$ is assumed to be drawn from the conditional distribution given the state,
\begin{align}
    P(a|s)
\end{align}
(this can also be interpreted as a stochastic driver policy, frequently denoted by $\pi(a|s)$).

The \textbf{main objective} of this work is to achieve an accurate estimation of the tail quantiles of $P(a|s)$, where by \emph{tail} quantiles we generally mean the (very) high/low quantiles. This is of major importance because in order to assess the safety of a driving scenario, it is often necessary to estimate the extreme ends of the driving behavior distribution, especially the high quantiles. By doing so, we can determine the maximum risk or most aggressive driving behavior of other vehicles that can be handled without causing collisions. Hence, accurately estimating the tail quantiles is essential to allow an AD vehicle to make well-informed decisions related to safety in various driving scenarios.

In order to fulfill the aforementioned objective, vehicle trajectories are extracted from the \gls{highd} dataset \cite{highDdataset}, and treated as \gls{iid} state-action pairs ${(s_0,a_0),(s_1,a_1),\ldots}$. From this data we then aim to learn the tail quantiles of the action distribution $P(a|s)$. This setup is similar to the standard behavioral cloning setting (i.e., imitation learning based on single-step supervised learning) \cite{osa2018algorithmic}, 
but with a focus on learning the action tail quantiles.
%

\section{APPROACH}
\label{sec:approach}

\subsection{Quantile Regression}


Consider the conditional probability distribution $P(a|s)$, where
a regression model aims to predict the value of $a$ given $s$. 
Often, one uses \gls{ls} regression, giving us an estimate of the conditional mean $E(a|s)$ \cite{wasserman2010statistics}.
However, the conditional mean may not always account for the potential variability or skewness in the distribution.
In contrast, the \gls{qr} method~\cite{koenker_quantile_2001} can account for those additional aspects of the conditional distribution.
This technique has been widely used in economics since it can estimate any quantile of the conditional distribution (including the median).


Analogously as \gls{ls} regression minimizes the squared-error loss function to predict the mean, \gls{qr} minimizes the quantile loss to predict a certain quantile.
Consider $a$ as the true action and $a^{pred}$ as the predicted action. The \emph{quantile loss} function of target quantile level $\alpha \in [0,1]$, also known as the \emph{\gls{tal} or pinball loss}, is denoted by
\begin{align}
L_{\alpha}(a,a^{pred}) &= \max\{\alpha \cdot (a - a^{pred}), (\alpha-1)\cdot(a - a^{pred})\},
\label{eq:tiltedabsoluteloss}
\end{align}

\begin{figure}[t] 
	\begin{center}
		\input{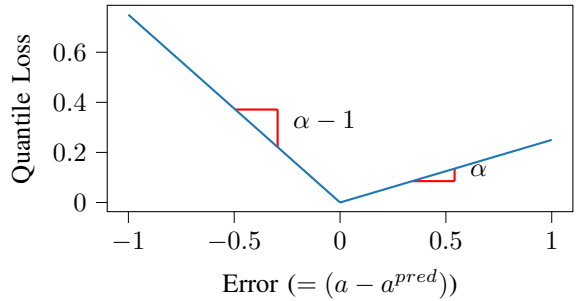}
		\caption{
		Quantile loss (Eq.~\eqref{eq:tiltedabsoluteloss}) for quantile level $\alpha$
		\cite{koenker_quantile_2001}.}
		 \label{fig:talf}
	\end{center}
\end{figure}

The intuition of the \gls{tal} is as follows: At higher quantiles, the positive errors (underpredictions) are more penalized, whereas at lower quantiles, the negative errors (overpredictions) are more penalized. 
This results in an asymmetrical loss line, as shown in Figure \ref{fig:talf}.
The quantile loss is unique for every quantile, as the quantile loss function is essentially a function of the quantile level $\alpha$ itself.

%
%
%
%
A key property of the \gls{tal} $L_\alpha$ is that, when averaging it over a dataset of $a$'s,
its \emph{minimizer for $a^{pred}$} is exactly the \emph{dataset's quantile for quantile level $\alpha$} \cite{koenker_quantile_2001}.
(This can easily be verified; the classic example being the median minimizing the (untilted) absolute error.)
This also allows to build consistent estimators for (conditional) quantiles \cite{si_autoregressive_2021}.
Additionally, it is computationally efficient and easily implemented. 
A shortcoming of \gls{qr} is that generically it is only defined for \gls{1d} settings.

We trained neural networks using the quantile loss function $L_\alpha$ (see Eq. \eqref{eq:tiltedabsoluteloss}) for each considered quantile level $\alpha$ (one network per level), to output the (tail) quantile values of interest of the conditional distribution $P(a|s)$, for a given set of state features (more details follow in \ref{sec:experiments}). 
The networks consist of 4 fully connected hidden layers with 64 neurons and \gls{relu} activation functions, and a linear output layer.

\subsection{Autoregressive Quantile Flows}


Besides the \gls{qr}, we build upon the \gls{aqf} method~\cite{si_autoregressive_2021}, which is a flexible class of latent variable-based, generative probabilistic models that can be used to accurately capture predictive aleatoric uncertainties. It applies to arbitrary dimensional variables, in contrast to \gls{qr}, which generically only applies to \gls{1d}.
We adapt the method to address the specific problem of driver modeling.
The general unconditional \gls{aqf} between the target variable $a \in \mathbb{R}^D$ and base variable $z \in \mathbb{R}^D$ ($z$ is a latent variable), $D\geq1$ is defined similarly to the autoregressive flow, i.e.,
\begin{equation}
    \label{eq:quantileflows}
	a_j = \tau(z_j;h_j), \qquad h_j = c_j(a_{<j}),
\end{equation}
where $\tau(z_j;h_j)$ is an invertible coupling transformer and $c_j(a_{<j})$ is the $j$-th conditioner on variables of the previous dimensions. 
However, unlike in standard \glspl{anf}~\cite{huang2018neural, padmanabha_solving_2020}, where negative log-likelihood is used as the loss function, the \glspl{aqf} learning objective~\cite{si_autoregressive_2021} ($L_{\acrshort{aqf}}$) is based on the quantile loss $L_{\alpha}$:
\begin{equation}
   L_{\acrshort{aqf}} = \frac{1}{n}\sum_{i=1}^{n}\sum_{j=1}^{D} L(a_{ij},a_{ij}^{pred}),
\end{equation}
with
\begin{align}
    L(a_{ij},a_{ij}^{pred}) 
    & = \int_{0}^{1} L_{\alpha}(a_{ij},a_{ij}^{pred}(\alpha)) d\alpha,
    \label{eq:lossintegral}
\end{align}
where $i$ ranges over the $n$ action samples, $j$ ranges over action dimensions $1, \ldots, D$, $a_{ij}^{pred}$ denotes the overall prediction, and $a_{ij}^{pred}(\alpha)$ the prediction (generated by the flow, as will be detailed below) for a specific quantile level $\alpha$.
In this formula, first, the quantile loss from Eq.~\eqref{eq:tiltedabsoluteloss}, for quantile level $\alpha$ (plugged in for $z$) and corresponding prediction $a_{ij}^{pred}(\alpha)$, is integrated over the quantile level $\alpha$'s range $[0, 1]$. 
Subsequently, the loss values from every dimension $j$ are summed, and the loss is averaged over $n$ samples. 
As the integral in Eq. \eqref{eq:lossintegral} is potentially intractable, Monte-Carlo sampling is used to approximate it, i.e.,
we sample $\alpha$ uniformly at random in $[0,1]$ \cite{si_autoregressive_2021}.
(This is opposed to \gls{anf}, where no such sampling is involved.)

We pass $\alpha_j$ together with $a_{<j}$ through the flow, and obtain the driver action of dimension $j$ (i.e., $a_j$) as an output. This process is also known as \emph{generation}.
One particular benefit of sampling the base variable from a uniform distribution is that in this case,
for any quantile level $\alpha$, the corresponding quantile of $P(a|s)$ is simply given by setting $z=\alpha$, i.e., by $\tau(\alpha)$, because the flow is the inverse of the CDF and thus represents the quantile function $Q$ (see \cite{si_autoregressive_2021} and Section~\ref{subsec:background}; this holds for the 1D case, and conditionally for higher dimensions).

\begin{figure}[t]
	\begin{center}
		\includegraphics[width=1\linewidth, height=0.14\textheight]{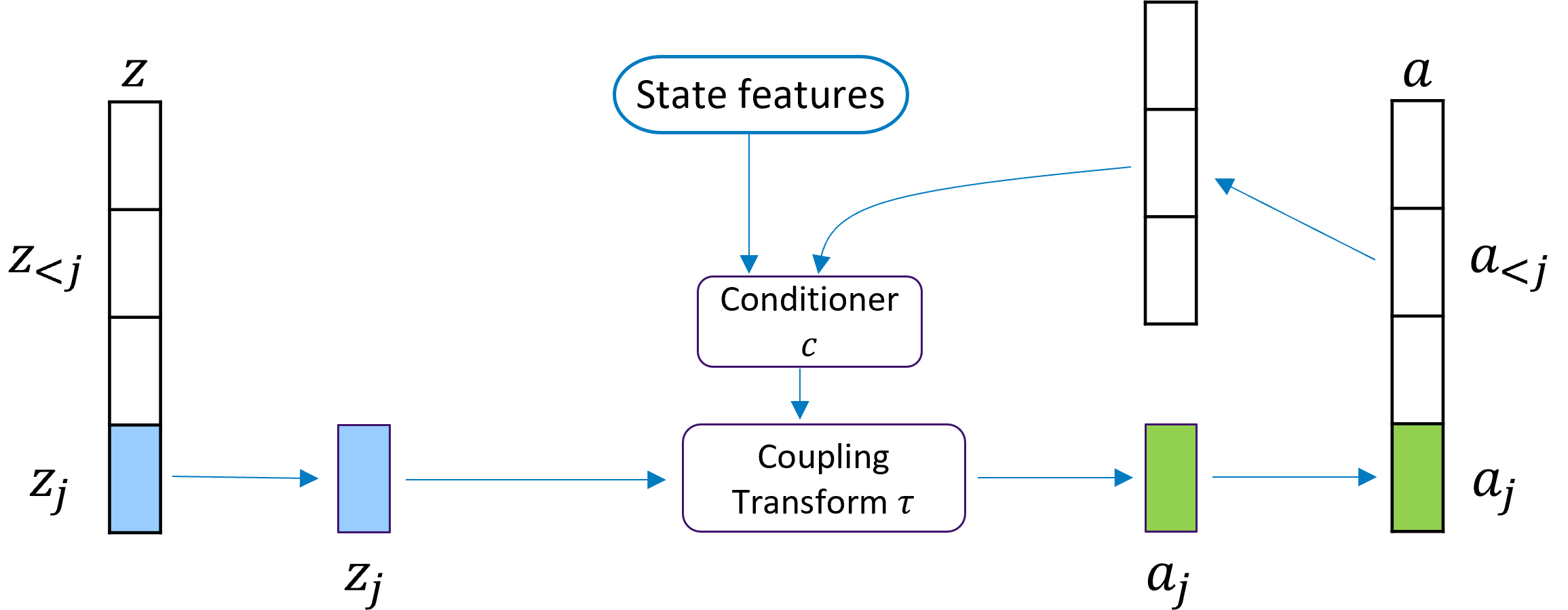}
		\caption{Structure of a conditional \gls{aqf} (generative direction).} 
		\label{fig:aqfbackward}
	\end{center}
\end{figure}

In order to include the state features in addition to the base variable $z$, we need a \emph{conditional} autoregressive quantile flow (a version of Eq.~\ref{eq:quantileflows}):
\begin{equation}
	\label{eq:conditionalquantileflows}
	a_j = \tau(z_j;h_j), \qquad h_j = c_j(a_{<j},g(s)),
\end{equation}
where $g(s)$ is the state feature mapping (in the case of $j=1$, the $a_{<j}$ is simply not considered as an argument of the function $c_j(\ldots)$). The overall structure of the conditional \gls{aqf} is shown in Figure \ref{fig:aqfbackward}.

The \glspl{aqf} have a similar functional form and coupling functions as building block to the standard \glspl{anf}. 
In this paper, we implement both \glspl{aqf} with affine coupling layers~
\cite{kobyzev_normalizing_2021} (\acrshort{aqf}-AFFINE) and \gls{nlsq} coupling layer~\cite{ziegler_latent_2019} (\acrshort{aqf}-\acrshort{nl}).

\section{EXPERIMENTS}
\label{sec:experiments}

To empirically evaluate the approaches introduced in Section \ref{sec:approach}, and compare their performance against the baselines, experiments are conducted using a dataset of driving scenarios. Here, we limit the setting to an exemplary highway scenario with only one modeled other vehicle in the surrounding of an AD ego vehicle. 
By focusing on modeling a single vehicle, we are able to gain deeper insights into the estimation of tail quantiles of the driving agent's actions in a more controlled setting, keeping a manageable level of complexity in the experiments.
We believe this can naturally be extended to multiple modeled vehicle agents.

In the first experiments (Section \ref{subsec:singlestepexp}), we want to predict the one-step acceleration value of a follower vehicle behind an ego vehicle as an action $a$ using the discussed approaches. We conduct both \gls{1d} and \gls{2d} experiments. In the 1D experiment, where only the longitudinal acceleration will be considered, we evaluate the discussed approaches: QR and AQFs against the baselines, including a \gls{dgp} \cite{ICML2018}, and \glspl{anf} \cite{papamakarios_masked_2017}. In the latter experiment, we employ the \glspl{aqf} to predict both the longitudinal and lateral acceleration values of the follower vehicle, and compare them against the baseline, namely the \glspl{anf}. It is worth noting that the \glspl{aqf} and \glspl{anf} employed in our study share the same architecture for both the 1D and the 2D cases, but they are trained separately.

In the second experiment (Section~\ref{subsec:multistepexp}), a multi-step rollout experiment is carried out in \gls{1d}, to test the proposed approaches in predicting the acceleration values of the follower vehicle for multiple time steps into the future, and to observe their performance under a simulated environment with different initial conditions.

\subsection{Dataset}
\label{subsec:datasets}
The dataset we utilize is taken from the ``Highway Drone Dataset'' (\gls{highd}) dataset \cite{highDdataset}. It covers multiple naturalistic vehicle trajectories recorded on six different German highway sections with \SI{25}{\hertz} sampling frequency using a drone. Each highway section is approximately \SI{420}{\meter} long and there are both light and heavy traffic trajectories. 
This dataset comprises diverse information about the driving actions of multiple drivers, including both longitudinal and lateral actions, as well as useful information about the surrounding traffic environment, such as \gls{dhw}, \gls{thw}, \gls{ttc}, and lane changes, which is well suited for our experiments.
In the \gls{1d} experiment, we use \gls{dhw}, \gls{thw}, \gls{ttc}, and velocities of both vehicles as state features. In the \gls{2d} case, the number of lanes on either side of the modeled agent is also included.

\subsection{Single-Step Experiments}
\label{subsec:singlestepexp}

\begin{table*}[!htb]
	\small\sffamily\centering%
	\setlength{\extrarowheight}{2pt}
	\caption{Quantile loss scores at different quantile levels for \gls{1d} experiment.}
	\label{tab:1dscore}
    \begin{adjustbox}{width=1.5\columnwidth,center}	
	\begin{tabular}{|*{11}{l|}}
		\hline
		\textbf{Methods} & \textbf{0.001} & \textbf{0.01} & \textbf{0.05} & \textbf{0.25} & \textbf{0.5} & \textbf{0.75} & \textbf{0.95} & \textbf{0.99} & \textbf{0.999} \\
		\hline
		\acrshort{dgp} (baseline) & 0.000313 &  0.001382 & 0.004374 & 0.012319 & 0.014853 & 0.011638 & 0.003977 & 0.001121 & 0.000178\\
		\hline
		\acrshort{qr} & \textbf{0.000218} & \textbf{0.001319} & \textbf{0.004283} & 0.011871 & 0.014571 & 0.011549 & \textbf{0.003902} & \textbf{0.001082} & \textbf{0.000153}\\
		\hline
		\acrshort{anf}-AFFINE (baseline) & 0.000256 & 0.001545 & 0.005318 & 0.013830 & 0.015790 & 0.012131 & 0.004755 & 0.001334 & 0.000182\\
		\hline
		\acrshort{anf}-\acrshort{nl} (baseline) & 0.000319 & 0.002026 & 0.005434 & 0.012397 & 0.014832 & 0.012135 & 0.004637 & 0.001450 & 0.000224\\
		\hline
		\acrshort{aqf}-AFFINE & 0.001679 & 0.002235 & 0.004524 & 0.012025 & 0.014598 & 0.011693 & 0.004109 & 0.001815 & 0.001259\\
		\hline
		\acrshort{aqf}-\acrshort{nl} & 0.000503 & 0.001374 & 0.004300 & \textbf{0.011850} & \textbf{0.014524} & \textbf{0.011519} & 0.003909 & 0.001286 & 0.000612\\
		\hline
	\end{tabular}
	\end{adjustbox}
\end{table*}

\begin{table*}[!htb]
	\small\sffamily\centering%
	\setlength{\extrarowheight}{2pt}
	\caption{Quantile loss scores at different quantile levels for 2D experiment.}
	\label{tab:2dscore}
    \begin{adjustbox}{width=1.5\columnwidth,center}	
	\begin{tabular}{|*{11}{l|}}
        \hline
        \textbf{Methods} & \textbf{0.001} & \textbf{0.01} & \textbf{0.05} & \textbf{0.25} & \textbf{0.5} & \textbf{0.75} & \textbf{0.95} & \textbf{0.99} & \textbf{0.999} \\
        \hline
		\acrshort{anf}-AFFINE (baseline) &  0.000865 & 0.002555 &	0.006964 & 0.017089 & 0.019947 & 0.015958 &	0.005912 &	\textbf{0.001827} &	\textbf{0.000435}\\
		\hline
		\acrshort{anf}-\acrshort{nl} (baseline) & \textbf{0.000465} &	\textbf{0.002274} &	\textbf{0.006829} &	0.016716 &	0.019605 & 0.015662 &	\textbf{0.005726} &	0.001973 &	0.000445\\
		\hline
		\acrshort{aqf}-AFFINE & 0.003746 &	0.004450 &	0.007342 &	0.016751 &	0.019546 & 0.015738 & 0.006083 & 0.003199 & 0.002516\\
		\hline
		\acrshort{aqf}-\acrshort{nl} & 0.001542 & 0.002663 & 0.006845 &	\textbf{0.016516} &	\textbf{0.019465} &	\textbf{0.015534} & 0.005772 & 0.002360 & 0.001503\\
		\hline
	\end{tabular}
	\end{adjustbox}
\end{table*}

In the single-step experiments, we aim to assess the performance of the discussed approaches (\gls{qr}, \gls{aqf}-AFFINE, \gls{aqf}-\gls{nl}) against the baselines (\gls{dgp}, \gls{anf}-AFFINE, \gls{anf}-\gls{nl}) by comparing their quantile loss values $L_{\alpha}(a,a^{pred})$ on the \gls{highd} test dataset at various quantile levels. The quantile loss values provide a straightforward intuition for evaluating the quality of the models, where a lower value indicates a better fit of the quantile. By examining the quantile loss values at different quantile levels, we can gain insight into the overall performance of the models in accurately estimating the quantiles of the one-step acceleration distribution.

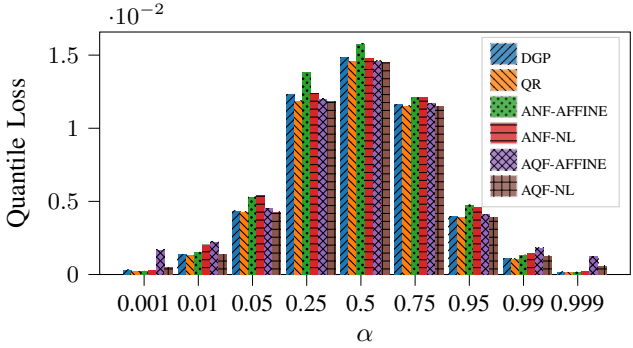
\begin{figure}
\centering
\pgfplotsset{compat=1.11,
        /pgfplots/ybar legend/.style={
        /pgfplots/legend image code/.code={%
        \draw[##1,/tikz/.cd,bar width=10pt,yshift=-0.2em,bar shift=0pt]
                plot coordinates {(0cm,0.8em)};},
},
}
\begin{tikzpicture}

\definecolor{crimson2143940}{RGB}{214,39,40}
\definecolor{darkgray176}{RGB}{176,176,176}
\definecolor{darkorange25512714}{RGB}{255,127,14}
\definecolor{forestgreen4416044}{RGB}{44,160,44}
\definecolor{lightgray204}{RGB}{204,204,204}
\definecolor{mediumpurple148103189}{RGB}{148,103,189}
\definecolor{sienna1408675}{RGB}{140,86,75}
\definecolor{steelblue31119180}{RGB}{31,119,180}

\begin{axis}[
legend cell align={left},
legend style={nodes={scale=0.7, transform shape,
font=\footnotesize}, fill opacity=0.8, draw opacity=1, text opacity=1, draw=lightgray204},
tick align=outside,
tick pos=left,
x grid style={darkgray176},
xlabel={$\alpha$},
xmin=-0.82, xmax=8.97,
xtick style={color=black},
xtick={0,1,2,3,4,5,6,7,8},
xticklabels={0.001,0.01,0.05,0.25,0.5,0.75,0.95,0.99,0.999},
y grid style={darkgray176},
ylabel={Quantile Loss},
ymin=0, ymax=0.0165793761,
ytick style={color=black},
tick label style = {font=\small},
width = \columnwidth,
height = 0.2\textheight
]
\draw[draw=none,fill=steelblue31119180,postaction={pattern=north east lines}] (axis cs:-0.375,0) rectangle (axis cs:-0.225,0.000312945);
\addlegendimage{ybar,ybar legend,draw=none,fill=steelblue31119180,postaction={pattern=north east lines}}
\addlegendentry{DGP}

\draw[draw=none,fill=steelblue31119180,postaction={pattern=north east lines}] (axis cs:0.625,0) rectangle (axis cs:0.775,0.001382407);
\draw[draw=none,fill=steelblue31119180,postaction={pattern=north east lines}] (axis cs:1.625,0) rectangle (axis cs:1.775,0.004373687);
\draw[draw=none,fill=steelblue31119180,postaction={pattern=north east lines}] (axis cs:2.625,0) rectangle (axis cs:2.775,0.012318899);
\draw[draw=none,fill=steelblue31119180,postaction={pattern=north east lines}] (axis cs:3.625,0) rectangle (axis cs:3.775,0.014853041);
\draw[draw=none,fill=steelblue31119180,postaction={pattern=north east lines}] (axis cs:4.625,0) rectangle (axis cs:4.775,0.011638222);
\draw[draw=none,fill=steelblue31119180,postaction={pattern=north east lines}] (axis cs:5.625,0) rectangle (axis cs:5.775,0.003976818);
\draw[draw=none,fill=steelblue31119180,postaction={pattern=north east lines}] (axis cs:6.625,0) rectangle (axis cs:6.775,0.001121169);
\draw[draw=none,fill=steelblue31119180,postaction={pattern=north east lines}] (axis cs:7.625,0) rectangle (axis cs:7.775,0.000177918);
\draw[draw=none,fill=darkorange25512714,postaction={pattern=north west lines}] (axis cs:-0.225,0) rectangle (axis cs:-0.075,0.000217626);
\addlegendimage{ybar,ybar legend,draw=none,fill=darkorange25512714,postaction={pattern=north west lines}}
\addlegendentry{QR}

\draw[draw=none,fill=darkorange25512714,postaction={pattern=north west lines}] (axis cs:0.775,0) rectangle (axis cs:0.925,0.001318543);
\draw[draw=none,fill=darkorange25512714,postaction={pattern=north west lines}] (axis cs:1.775,0) rectangle (axis cs:1.925,0.004283488);
\draw[draw=none,fill=darkorange25512714,postaction={pattern=north west lines}] (axis cs:2.775,0) rectangle (axis cs:2.925,0.011871061);
\draw[draw=none,fill=darkorange25512714,postaction={pattern=north west lines}] (axis cs:3.775,0) rectangle (axis cs:3.925,0.014571);
\draw[draw=none,fill=darkorange25512714,postaction={pattern=north west lines}] (axis cs:4.775,0) rectangle (axis cs:4.925,0.01154907);
\draw[draw=none,fill=darkorange25512714,postaction={pattern=north west lines}] (axis cs:5.775,0) rectangle (axis cs:5.925,0.003902226);
\draw[draw=none,fill=darkorange25512714,postaction={pattern=north west lines}] (axis cs:6.775,0) rectangle (axis cs:6.925,0.001081901);
\draw[draw=none,fill=darkorange25512714,postaction={pattern=north west lines}] (axis cs:7.775,0) rectangle (axis cs:7.925,0.000153183);
\draw[draw=none,fill=forestgreen4416044,postaction={pattern=crosshatch dots}] (axis cs:-0.075,0) rectangle (axis cs:0.075,0.000256141);
\addlegendimage{ybar,ybar legend,draw=none,fill=forestgreen4416044,postaction={pattern=crosshatch dots}}
\addlegendentry{ANF-AFFINE}

\draw[draw=none,fill=forestgreen4416044,postaction={pattern=crosshatch dots}] (axis cs:0.925,0) rectangle (axis cs:1.075,0.001545108);
\draw[draw=none,fill=forestgreen4416044,postaction={pattern=crosshatch dots}] (axis cs:1.925,0) rectangle (axis cs:2.075,0.00531804);
\draw[draw=none,fill=forestgreen4416044,postaction={pattern=crosshatch dots}] (axis cs:2.925,0) rectangle (axis cs:3.075,0.013829959);
\draw[draw=none,fill=forestgreen4416044,postaction={pattern=crosshatch dots}] (axis cs:3.925,0) rectangle (axis cs:4.075,0.015789882);
\draw[draw=none,fill=forestgreen4416044,postaction={pattern=crosshatch dots}] (axis cs:4.925,0) rectangle (axis cs:5.075,0.01213062);
\draw[draw=none,fill=forestgreen4416044,postaction={pattern=crosshatch dots}] (axis cs:5.925,0) rectangle (axis cs:6.075,0.004754792);
\draw[draw=none,fill=forestgreen4416044,postaction={pattern=crosshatch dots}] (axis cs:6.925,0) rectangle (axis cs:7.075,0.001334461);
\draw[draw=none,fill=forestgreen4416044,postaction={pattern=crosshatch dots}] (axis cs:7.925,0) rectangle (axis cs:8.075,0.000181896);
\draw[draw=none,fill=crimson2143940,postaction={pattern=horizontal lines}] (axis cs:0.075,0) rectangle (axis cs:0.225,0.000318512);
\addlegendimage{ybar,ybar legend,draw=none,fill=crimson2143940,postaction={pattern=horizontal lines}}
\addlegendentry{ANF-NL}

\draw[draw=none,fill=crimson2143940,postaction={pattern=horizontal lines}] (axis cs:1.075,0) rectangle (axis cs:1.225,0.002026129);
\draw[draw=none,fill=crimson2143940,postaction={pattern=horizontal lines}] (axis cs:2.075,0) rectangle (axis cs:2.225,0.005433838);
\draw[draw=none,fill=crimson2143940,postaction={pattern=horizontal lines}] (axis cs:3.075,0) rectangle (axis cs:3.225,0.012397327);
\draw[draw=none,fill=crimson2143940,postaction={pattern=horizontal lines}] (axis cs:4.075,0) rectangle (axis cs:4.225,0.014831762);
\draw[draw=none,fill=crimson2143940,postaction={pattern=horizontal lines}] (axis cs:5.075,0) rectangle (axis cs:5.225,0.012135394);
\draw[draw=none,fill=crimson2143940,postaction={pattern=horizontal lines}] (axis cs:6.075,0) rectangle (axis cs:6.225,0.004636943);
\draw[draw=none,fill=crimson2143940,postaction={pattern=horizontal lines}] (axis cs:7.075,0) rectangle (axis cs:7.225,0.001450113);
\draw[draw=none,fill=crimson2143940,postaction={pattern=horizontal lines}] (axis cs:8.075,0) rectangle (axis cs:8.225,0.00022443);
\draw[draw=none,fill=mediumpurple148103189,postaction={pattern=crosshatch}] (axis cs:0.225,0) rectangle (axis cs:0.375,0.001679106);
\addlegendimage{ybar,ybar legend,draw=none,fill=mediumpurple148103189,postaction={pattern=crosshatch}}
\addlegendentry{AQF-AFFINE}

\draw[draw=none,fill=mediumpurple148103189,postaction={pattern=crosshatch}] (axis cs:1.225,0) rectangle (axis cs:1.375,0.002234821);
\draw[draw=none,fill=mediumpurple148103189,postaction={pattern=crosshatch}] (axis cs:2.225,0) rectangle (axis cs:2.375,0.00452419);
\draw[draw=none,fill=mediumpurple148103189,postaction={pattern=crosshatch}] (axis cs:3.225,0) rectangle (axis cs:3.375,0.012025115);
\draw[draw=none,fill=mediumpurple148103189,postaction={pattern=crosshatch}] (axis cs:4.225,0) rectangle (axis cs:4.375,0.014598469);
\draw[draw=none,fill=mediumpurple148103189,postaction={pattern=crosshatch}] (axis cs:5.225,0) rectangle (axis cs:5.375,0.01169314);
\draw[draw=none,fill=mediumpurple148103189,postaction={pattern=crosshatch}] (axis cs:6.225,0) rectangle (axis cs:6.375,0.004108929);
\draw[draw=none,fill=mediumpurple148103189,postaction={pattern=crosshatch}] (axis cs:7.225,0) rectangle (axis cs:7.375,0.001814902);
\draw[draw=none,fill=mediumpurple148103189,postaction={pattern=crosshatch}] (axis cs:8.225,0) rectangle (axis cs:8.375,0.001258538);
\draw[draw=none,fill=sienna1408675,postaction={pattern=grid}] (axis cs:0.375,0) rectangle (axis cs:0.525,0.000502835);
\addlegendimage{ybar,ybar legend,draw=none,fill=sienna1408675,postaction={pattern=grid}}
\addlegendentry{AQF-NL}

\draw[draw=none,fill=sienna1408675,postaction={pattern=grid}] (axis cs:1.375,0) rectangle (axis cs:1.525,0.001373981);
\draw[draw=none,fill=sienna1408675,postaction={pattern=grid}] (axis cs:2.375,0) rectangle (axis cs:2.525,0.004299616);
\draw[draw=none,fill=sienna1408675,postaction={pattern=grid}] (axis cs:3.375,0) rectangle (axis cs:3.525,0.011850363);
\draw[draw=none,fill=sienna1408675,postaction={pattern=grid}] (axis cs:4.375,0) rectangle (axis cs:4.525,0.014524128);
\draw[draw=none,fill=sienna1408675,postaction={pattern=grid}] (axis cs:5.375,0) rectangle (axis cs:5.525,0.011518706);
\draw[draw=none,fill=sienna1408675,postaction={pattern=grid}] (axis cs:6.375,0) rectangle (axis cs:6.525,0.003909175);
\draw[draw=none,fill=sienna1408675,postaction={pattern=grid}] (axis cs:7.375,0) rectangle (axis cs:7.525,0.001286063);
\draw[draw=none,fill=sienna1408675,postaction={pattern=grid}] (axis cs:8.375,0) rectangle (axis cs:8.525,0.000611576);
\end{axis}

\end{tikzpicture}
\caption{Quantile loss at different quantile levels for \acrshort{1d} case.}
\label{fig:qlossfeatureB}
\end{figure}

Table \ref{tab:1dscore} shows the quantile loss score of each method in the \gls{1d} single-step experiment for nine quantile levels 0.001, 0.01, 0.05, 0.25, 0.50, 0.75, 0.95, 0.99, 0.999 (while the focus of this work is on the high/low tail quantiles, we also included some quantile levels around the median to ensure reasonable performance also for these parts of the distribution). The best (lowest) quantile loss value for each quantile level is indicated in bold. 
%
Additionally, in order to facilitate a clearer comparison of the quantile loss results across different approaches, a bar chart displaying the quantile loss scores for each feature set are also included (see Figure \ref{fig:qlossfeatureB}).


An analysis of Table \ref{tab:1dscore} reveals that the \gls{qr} method estimates the tail quantiles of the conditional acceleration distribution better than other models, as evidenced by its lowest quantile loss scores. It also exhibits satisfactory performance in estimating quantiles near the center of the distribution (0.25, 0.50, 0.75), with the second lowest quantile loss scores, slightly behind \gls{aqf}-\gls{nl}. This observation suggests that the \gls{qr} approach performs well in capturing the overall pattern of the conditional acceleration distribution of the follower vehicle, particularly when it comes to rare driving behavior.


On the other hand, the \gls{aqf}-\gls{nl} approach achieves relatively good quantile loss scores for most quantile levels, except for the extreme quantile levels (0.001, 0.999). In contrast,  the \gls{aqf}-AFFINE approach is less effective and yields quantile loss values that are more than twice as high as those of the former method at the tail quantiles. This is attributed to the limited expressiveness of the affine coupling function used in the flow, in comparison to the nonlinear coupling function employed in \gls{aqf}-\gls{nl}. While both \gls{anf}-AFFINE and \gls{anf}-\gls{nl} perform better than the \gls{aqf} methods at the extreme quantile levels 0.001 and 0.999, their quantile fitting performance deteriorates as the quantile level approaches the median.

\begin{figure}
\centering
\pgfplotsset{compat=1.11,
        /pgfplots/ybar legend/.style={
        /pgfplots/legend image code/.code={%
        \draw[##1,/tikz/.cd,bar width=10pt,yshift=-0.2em,bar shift=0pt]
                plot coordinates {(0cm,0.8em)};},
},
}
\begin{tikzpicture}

\definecolor{crimson2143940}{RGB}{214,39,40}
\definecolor{darkgray176}{RGB}{176,176,176}
\definecolor{darkorange25512714}{RGB}{255,127,14}
\definecolor{forestgreen4416044}{RGB}{44,160,44}
\definecolor{lightgray204}{RGB}{204,204,204}
\definecolor{mediumpurple148103189}{RGB}{148,103,189}
\definecolor{sienna1408675}{RGB}{140,86,75}
\definecolor{steelblue31119180}{RGB}{31,119,180}

\begin{axis}[
legend cell align={left},
legend style={fill opacity=0.8, draw opacity=1, text opacity=1, draw=lightgray204,
font=\footnotesize},
tick align=outside,
tick pos=left,
x grid style={darkgray176},
xlabel={$\alpha$},
xmin=-0.505, xmax=8.955,
xtick style={color=black},
xtick={0,1,2,3,4,5,6,7,8},
xticklabels={0.001,0.01,0.05,0.25,0.5,0.75,0.95,0.99,0.999},
y grid style={darkgray176},
ylabel={Quantile Loss},
ymin=0, ymax=0.0209445894,
ytick style={color=black},
tick label style = {font=\small},
width = \columnwidth,
height = 0.2\textheight
]
\draw[draw=none,fill=forestgreen4416044,postaction={pattern=crosshatch dots}] (axis cs:-0.075,0) rectangle (axis cs:0.075,0.000864541);
\addlegendimage{ybar,ybar legend,draw=none,fill=forestgreen4416044,postaction={pattern=crosshatch dots}}
\addlegendentry{ANF-AFFINE}

\draw[draw=none,fill=forestgreen4416044,postaction={pattern=crosshatch dots}] (axis cs:0.925,0) rectangle (axis cs:1.075,0.002554906);
\draw[draw=none,fill=forestgreen4416044,postaction={pattern=crosshatch dots}] (axis cs:1.925,0) rectangle (axis cs:2.075,0.00696429);
\draw[draw=none,fill=forestgreen4416044,postaction={pattern=crosshatch dots}] (axis cs:2.925,0) rectangle (axis cs:3.075,0.017088728);
\draw[draw=none,fill=forestgreen4416044,postaction={pattern=crosshatch dots}] (axis cs:3.925,0) rectangle (axis cs:4.075,0.019947228);
\draw[draw=none,fill=forestgreen4416044,postaction={pattern=crosshatch dots}] (axis cs:4.925,0) rectangle (axis cs:5.075,0.015958069);
\draw[draw=none,fill=forestgreen4416044,postaction={pattern=crosshatch dots}] (axis cs:5.925,0) rectangle (axis cs:6.075,0.005911909);
\draw[draw=none,fill=forestgreen4416044,postaction={pattern=crosshatch dots}] (axis cs:6.925,0) rectangle (axis cs:7.075,0.001827395);
\draw[draw=none,fill=forestgreen4416044,postaction={pattern=crosshatch dots}] (axis cs:7.925,0) rectangle (axis cs:8.075,0.000435328);
\draw[draw=none,fill=crimson2143940,postaction={pattern=horizontal lines}] (axis cs:0.075,0) rectangle (axis cs:0.225,0.000465075);
\addlegendimage{ybar,ybar legend,draw=none,fill=crimson2143940,postaction={pattern=horizontal lines}}
\addlegendentry{ANF-NL}

\draw[draw=none,fill=crimson2143940,postaction={pattern=horizontal lines}] (axis cs:1.075,0) rectangle (axis cs:1.225,0.00227394);
\draw[draw=none,fill=crimson2143940,postaction={pattern=horizontal lines}] (axis cs:2.075,0) rectangle (axis cs:2.225,0.006828588);
\draw[draw=none,fill=crimson2143940,postaction={pattern=horizontal lines}] (axis cs:3.075,0) rectangle (axis cs:3.225,0.016716182);
\draw[draw=none,fill=crimson2143940,postaction={pattern=horizontal lines}] (axis cs:4.075,0) rectangle (axis cs:4.225,0.019605488);
\draw[draw=none,fill=crimson2143940,postaction={pattern=horizontal lines}] (axis cs:5.075,0) rectangle (axis cs:5.225,0.015661885);
\draw[draw=none,fill=crimson2143940,postaction={pattern=horizontal lines}] (axis cs:6.075,0) rectangle (axis cs:6.225,0.005725778);
\draw[draw=none,fill=crimson2143940,postaction={pattern=horizontal lines}] (axis cs:7.075,0) rectangle (axis cs:7.225,0.001972924);
\draw[draw=none,fill=crimson2143940,postaction={pattern=horizontal lines}] (axis cs:8.075,0) rectangle (axis cs:8.225,0.000444599);
\draw[draw=none,fill=mediumpurple148103189,postaction={pattern=crosshatch}] (axis cs:0.225,0) rectangle (axis cs:0.375,0.003746175);
\addlegendimage{ybar,ybar legend,draw=none,fill=mediumpurple148103189,postaction={pattern=crosshatch}}
\addlegendentry{AQF-AFFINE}

\draw[draw=none,fill=mediumpurple148103189,postaction={pattern=crosshatch}] (axis cs:1.225,0) rectangle (axis cs:1.375,0.004449973);
\draw[draw=none,fill=mediumpurple148103189,postaction={pattern=crosshatch}] (axis cs:2.225,0) rectangle (axis cs:2.375,0.007342801);
\draw[draw=none,fill=mediumpurple148103189,postaction={pattern=crosshatch}] (axis cs:3.225,0) rectangle (axis cs:3.375,0.016751123);
\draw[draw=none,fill=mediumpurple148103189,postaction={pattern=crosshatch}] (axis cs:4.225,0) rectangle (axis cs:4.375,0.019545897);
\draw[draw=none,fill=mediumpurple148103189,postaction={pattern=crosshatch}] (axis cs:5.225,0) rectangle (axis cs:5.375,0.015738076);
\draw[draw=none,fill=mediumpurple148103189,postaction={pattern=crosshatch}] (axis cs:6.225,0) rectangle (axis cs:6.375,0.006083098);
\draw[draw=none,fill=mediumpurple148103189,postaction={pattern=crosshatch}] (axis cs:7.225,0) rectangle (axis cs:7.375,0.003199242);
\draw[draw=none,fill=mediumpurple148103189,postaction={pattern=crosshatch}] (axis cs:8.225,0) rectangle (axis cs:8.375,0.002516209);
\draw[draw=none,fill=sienna1408675,postaction={pattern=grid}] (axis cs:0.375,0) rectangle (axis cs:0.525,0.001542438);
\addlegendimage{ybar,ybar legend,draw=none,fill=sienna1408675,postaction={pattern=grid}}
\addlegendentry{AQF-NL}

\draw[draw=none,fill=sienna1408675,postaction={pattern=grid}] (axis cs:1.375,0) rectangle (axis cs:1.525,0.002662723);
\draw[draw=none,fill=sienna1408675,postaction={pattern=grid}] (axis cs:2.375,0) rectangle (axis cs:2.525,0.006845371);
\draw[draw=none,fill=sienna1408675,postaction={pattern=grid}] (axis cs:3.375,0) rectangle (axis cs:3.525,0.016516447);
\draw[draw=none,fill=sienna1408675,postaction={pattern=grid}] (axis cs:4.375,0) rectangle (axis cs:4.525,0.019464678);
\draw[draw=none,fill=sienna1408675,postaction={pattern=grid}] (axis cs:5.375,0) rectangle (axis cs:5.525,0.015533712);
\draw[draw=none,fill=sienna1408675,postaction={pattern=grid}] (axis cs:6.375,0) rectangle (axis cs:6.525,0.005771859);
\draw[draw=none,fill=sienna1408675,postaction={pattern=grid}] (axis cs:7.375,0) rectangle (axis cs:7.525,0.002359509);
\draw[draw=none,fill=sienna1408675,postaction={pattern=grid}] (axis cs:8.375,0) rectangle (axis cs:8.525,0.001502874);
\end{axis}

\end{tikzpicture}
\caption{Quantile loss at different quantile levels for \acrshort{2d} case.}
\label{fig:qlossfeatureD}
\end{figure}
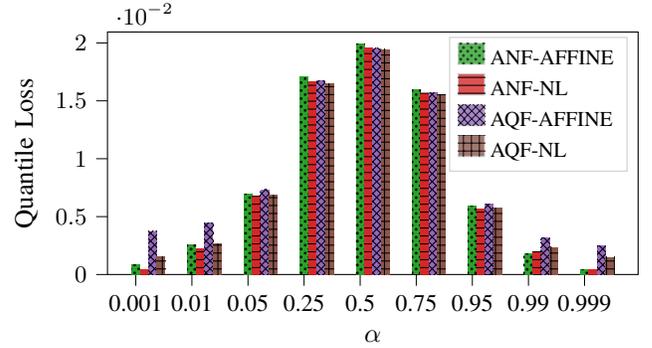

Table \ref{tab:2dscore} and Figure \ref{fig:qlossfeatureD} present the quantile loss scores of the \gls{anf} and \gls{aqf} methods in the \gls{2d} single-step experiment. The quantile loss is computed individually for each dimension, and then the average is taken to obtain the overall quantile loss value. Similar to Table \ref{tab:1dscore}, the best quantile loss value for each quantile level is highlighted in bold.


As one can clearly see from the table, while the \glspl{aqf} demonstrate comparable quantile loss scores to the \glspl{anf} in the quantile level range between 0.05 and 0.95, they consistently underperform at extreme quantile levels (0.001, 0.01, 0.99, 0.999). This difference is especially prominent for \gls{aqf}-AFFINE, which exhibits the highest quantile loss values among all methods due to the limitations of its affine coupling function. As mentioned in the introduction chapter, the \gls{aqf} approach is still in its early stages of development. One reason could be that, in the Monte-Carlo integration of Eq.~\eqref{eq:lossintegral} during training, for very high quantile levels $z$, there are only a few or even no quantile levels sampled above $z$, leading to non-optimal fits for these $z$.
Therefore, there is room for further refinement of this method, which provides a direction for future research.

\subsection{Multi-step Rollout Experiment}
\label{subsec:multistepexp}
In this section, the multi-step rollout observations in a \gls{1d} vehicle-following scenario are shown and discussed. 
%
%
%
We create a \gls{1d} simulation environment for the vehicle-following scenario with the highway-env library \cite{highway-env}. 
We filter for diverse ground truth driving scenarios of one vehicle following another (similar to \ref{fig:uncertaintrajectoryillustration}) from the \gls{highd} dataset and extract the joint trajectories. 
The agent models we trained are plugged into the simulation and output the acceleration value for the follower vehicle for every time step, while the front vehicle just executes its ground truth trajectory, i.e., performs log replay. 
The initial simulation state of all vehicles is set to match the true trajectory's initial position and velocity. 

\begin{figure}
\centering
    \input{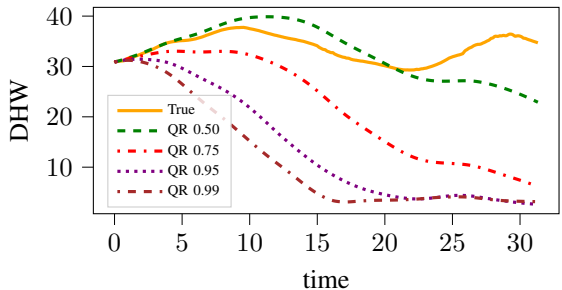}
\caption{DHW vs time plot of modeled agent by \acrshort{qr}.}
\label{fig:dhwvstimeqr}
\end{figure}

\begin{figure}
\centering
    \input{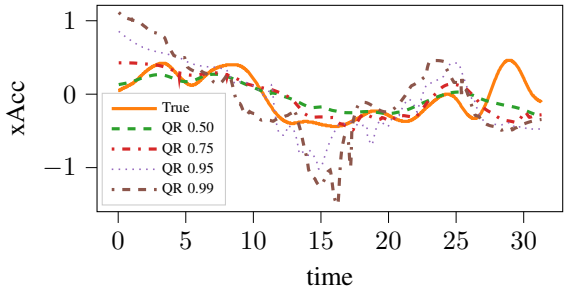}
\caption{Acceleration vs time plot of modeled agent by \acrshort{qr}.}
\label{fig:xaccvstimeqr}
\end{figure}

To gain a deeper understanding of the observations, several rollout trajectories of the QR model at quantile levels $\alpha = 0.50, 0.75, 0.95, 0.99$ are visually illustrated in Figures \ref{fig:dhwvstimeqr} and \ref{fig:xaccvstimeqr}.
This shows how \gls{dhw} decreases and aggressiveness of acceleration increases, with growing quantile levels, as we will discuss in Section \ref{subsec:diversebehavior}. (Note that if the estimated action distribution is too broad, then this can require overcautious AD maneuvering; a detailed investigation of this aspect is left to future work.)

While the multi-step rollout can provide some insights into the performance of the models beyond the single-step setting, it should not be considered a definitive or comprehensive evaluation in and of itself. This is particularly true since the models were trained as single-step predictors in an open-loop setting, so they may not be optimized or well-suited for predicting multiple steps into the future. Therefore, the multi-step rollout can be considered as a first impression or rough estimate of the models' predictive capabilities in a multi-step setting, rather than a definitive evaluation. 



\section{DISCUSSION}
\label{sec:discussion}
%
Beyond the analysis already discussed in the experiments, let us highlight some additional insights into the methods and experiments.

\subsection{Relationship between Single-Step Quantile Estimation and Multi-step Rollouts}

In the experiment, the trajectories from the \gls{highd} dataset are treated as \gls{iid} state-action pairs for training the supervised learning models. However, in a multi-step rollout, it becomes apparent that the action taken at one time step indirectly affects the subsequent state, violating the \gls{iid} assumption. This becomes even more critical when predicting the high quantiles, as the accumulation of aggressive actions can lead to very rare or non-recoverable states. This issue could also be regarded as covariate shift, in which the distribution of the states shifts gradually between the training environment and live environment over time \cite{quinonero2008dataset}, and it is similar to, but rather stronger than, the compounding error \cite{_pdf_????-1} limitation in an open-loop behavioral cloning method. 

To alleviate this issue, we use data balancing techniques, especially oversampling \cite{6542392}, which involves randomly duplicating the instances of the rare datapoints to balance it with the majority datapoints, so that they do not get overwhelmed by the other predominant datapoints during training. However, caution must be taken to ensure that the rebalancing is applied only to the training dataset and not the validation or test sets to avoid biased results. Alternatively, collecting more diverse training data or designing models that consider multiple past time steps for action prediction could help mitigate this issue.

Aside from that, the relationship between a single-step quantile and the quantile reached at the end of the multi-step rollout is not fully clear. It should be noted that maintaining a constant output of a high quantile level acceleration value for each time step in a rollout does not result in the same quantile level for the state reached at the end of the rollout; rather, it becomes higher. Addressing this issue requires a comprehensive understanding of power series analysis in the far tail. Existing calculations of power series expansion in the far tail discussed in \cite{_pdf_????-2} assume independence among time steps, which is not applicable in our multi-step rollout experiment. Consequently, the formula used in the literature cannot accurately calculate the final quantile of the conditional action distribution. Further research is needed to develop an exact, closed-form calculation for multi-step quantiles based on single-step quantiles.

\subsection{Diverse Vehicle-Following Behaviors}
\label{subsec:diversebehavior}
In our multi-step simulations of vehicle-following scenarios, we observe a wide range of tailing behaviors, from conservative to aggressive, demonstrating the ability of our approach to realistically capture the diversity of real-world driving. In particular, the plot in Figure \ref{fig:dhwvstimeqr} shows that the \gls{qr} method exhibits a smooth \emph{tailgating behavior} at $\alpha=0.99$, particularly when the follower vehicle approaches very close to the front vehicle. The method maintains a distance gap that was similar to the \gls{qr} at $\alpha=0.95$. This behavior closely resembles the tailgating behavior of an aggressive human driver. This indicates that the \gls{qr} approach has the capability to generate diverse and realistic range tailgating behaviors of vehicles. However, it is important to note that this is just one example, and in different traffic scenarios, the method may still produce undesirable acceleration values. Further investigation and analysis are required to comprehensively assess the performance and generalizability of our approach across a wider range of traffic scenarios and driving conditions. 

\subsection{Training Process of \gls{qr} for Several Quantiles}
In the \gls{1d} single-step experiment, the \gls{qr} method demonstrates superior performance compared to the baselines in accurately estimating the tail quantiles. However, it is important to highlight that QR involves training separate models for each quantile level. This means that individual models need to be trained to estimate different quantiles of the conditional action distribution. In contrast, the \gls{dgp}, \glspl{anf}, and \glspl{aqf} only require a single training process. These methods directly model the conditional action distribution without the need for explicit quantile estimation. This distinction in training requirements between \gls{qr} and the other approaches should be taken into consideration when choosing an appropriate method for quantile estimation.

\section{CONCLUSIONS}
\label{sec:conclusion}
In this paper, we took steps towards more accurately
learning the diversity and tail quantiles of driver action distributions given real-world driving trajectory data. 
Driver models were built using quantile regression and probabilistic autoregressive quantile flows, based on the single-step behavioral cloning framework. 
In the experiments of vehicle-following highway scenarios, we evaluated how these approaches can estimate the conditional acceleration distribution's quantiles, showing strengths and limitations of the individual approaches.
Additional experimental insights on mitigating covariate shift and on the diversity of behaviors were discussed.
Overall, this shows where the potential of these approaches lies towards flexibly estimating the long tails of realistic driver distributions and, in turn, improving the safety of autonomous driving.

It needs to be noted though, that the experiments and conclusions are limited and only constitute first steps.
For a broader understanding, further expriments are needed with
a deeper investigation of multi-step, multi-vehicle and lane-changing settings,
as well as more different traffic scenarios and bigger data sets.
Also the deployment and assessment of our models in concrete AD safety tests is future work.
But we believe that the methodology
presented in this work can be built on to 
also cover such more complex scenarios,
and see this as potential future work.




\addtolength{\textheight}{0cm}   






\section*{ACKNOWLEDGMENT}

The authors would like to thank Christoph-Nikolas Straehle for insightful discussions on normalizing flows, and the anonymous reviewers for their constructive feedback.


\bibliographystyle{IEEEtran}
\bibliography{citations}

\end{document}